\documentclass{llncs}

\usepackage{xspace}
\usepackage{todonotes}
\usepackage[nolist]{acronym}
\usepackage{hyperref}

\newcommand{\sparql}{\textsc{Sparql}\xspace}

\newcommand{\kbqa}{\textsc{Kbqa}\xspace}
\newcommand{\lcquad}{\textsc{Lc-Quad}\xspace}
\newcommand{\qald}{\textsc{Qald}\xspace}

\newcommand{\bioasq}{\textsc{BioAsq}\xspace}
\newcommand{\wqsp}{\textsc{WebQuestionsSp}\xspace}

\newcommand{\lod}{\textsc{Lod}\xspace}
\newcommand{\simpleq}{\textsc{SimpleQuestions}\xspace}
\newcommand{\complexq}{\textsc{ComplexQuestions}\xspace}

\begin{acronym}[UML]
	\acro{AOS}{Agricultural Ontology Services}
	\acro{AGRIS}{Agricultural Science and Technology}
	\acro{API}{Application Programming Interface}
	\acro{A2KB}{Annotation to Knowledge Base}
	\acro{BPSO}{Binary Particle-Swarm Optimization}
	\acro{BPMLOD}{Best Practices for Multilingual Linked Open Data}
	\acro{BFS}{Breadth-First-Search}
	\acro{CBD}{Concise Bounded Description}
	\acro{COG}{Content Oriented Guidelines}
	\acro{CSV}{Comma-Separated Values}
	\acro{CCR}{cross-document co-reference}
	\acro{DPSO}{Deterministic Particle-Swarm Optimization}
	\acro{DALY}{Disability Adjusted Life Year}
	\acro{D2KB}{Disambiguation to Knowledge Base}

	\acro{ER}{Entity Resolution}
	\acro{EM}{Expectation Maximization}
	\acro{EL}{Entity Linking}
	\acro{FAO}{Food and Agriculture Organization of the United Nations}
	\acro{GIS}{Geographic Information Systems}
	\acro{GHO}{Global Health Observatory}
	\acro{HDI}{Human Development Index}
	\acro{ICT}{Information and communication technologies}
    \acro{KB}{Knowledge Base}
	\acro{LR}  {Language Resource}
	\acro{LD}  {Linked Data}
	\acro{LLOD}  {Linguistic Linked Open Data}
	\acro{LIMES}{LInk discovery framework for MEtric Spaces}
	\acro{LS}  {Link Specifications}
	\acro{LDIF}{Linked Data Integration Framework}
	\acro{LGD} {LinkedGeoData}
	\acro{LOD} {Linked Open Data}
    \acro{LSTM} {Long-Short Term Memory Layers}
	\acro{MSE}{Mean Squared Error}
	\acro{MWE}{Multiword Expressions}
	\acro{NIF}{NLP Interchange Format}
	\acro{NIF4OGGD}{NLP Interchange Format for Open German Governmental Data}
	\acro{NLP}{Natural Language Processing}
	\acro{NER}{Named Entity Recognition}
	\acro{NED}{Named Entity Disambiguation}
	\acro{NEL}{Named Entity Linking}
	\acro{NN}{Neural Network}
    \acro{NLG}{Natural Language Generation}
	\acro{OSM}{OpenStreetMap}
	\acro{OWL}{Web Ontology Language}
	\acro{PFM}{Pseudo-F-Measures}
	\acro{PSO}{Particle-Swarm Optimization}
	\acro{QA}{Question Answering}
	\acro{RDF}{Resource Description Framework}
    \acro{REG}{Referring Expression Generation}
	\acro{SKOS}{Simple Knowledge Organization System}
	\acro{SPARQL}{SPARQL Protocol and RDF Query Language}
	\acro{SRL}{Statistical Relational Learning}
	\acro{SF}{surface forms}
    \acro{SW}{Semantic Web}
	\acro{UML}{Unified Modeling Language}
	\acro{WHO}{World Health Organization}
	\acro{WKT}{Well-Known Text}
	\acro{W3C}{World Wide Web Consortium}
	\acro{YPLL}{Years of Potential Life Lost}
\end{acronym}  
\begin{document}
\title{Where is Linked Data\\ in Question Answering over Linked Data?}
\titlerunning{Where is Linked Data in Question Answering over Linked Data?}  % abbreviated title (for running head)
\author{Tommaso Soru\inst{1}\inst{2} \and Edgard Marx\inst{1}\inst{3} \and Andr\'e Valdestilhas\inst{2} \and\\Diego Moussallem\inst{2} \and Gustavo Publio\inst{2} \and Muhammad Saleem\inst{2}}
\authorrunning{Tommaso Soru et al.} % abbreviated author list (for running head)
% \tocauthor{}
%
\institute{
Liber AI Research, London, UK \and
AKSW, University of Leipzig, Germany \and
HTWK Leipzig, Germany \\
\email{tom@tommaso-soru.it}}

\maketitle              % typeset the title of the contribution

\begin{abstract}
We argue that ``Question Answering with Knowledge Base'' and ``Question Answering over Linked Data'' are currently two instances of the same problem, despite one explicitly declares to deal with Linked Data.
We point out the lack of existing methods to evaluate question answering on datasets which exploit external links to the rest of the cloud or share common schema.
To this end, we propose the creation of new evaluation settings to leverage the advantages of the Semantic Web to achieve AI-complete question answering.
\end{abstract}

\section{Introduction}

Question Answering with Knowledge Base (\kbqa) parses a natural-language question and returns an appropriate answer that can be found in a \ac{KB}.
Currently, one of the most exciting scenarios for \ac{QA} is the Web of Data, a fast-growing distributed cloud of interlinked \ac{KB}s which comprises more than 100 billions of edges~\cite{mccrae2018lod}.
Similarly, Question Answering over Linked Data (\qald) is a research field aimed at transforming utterances into \sparql queries which can be executed towards the Linked Open Data (\lod) cloud~\cite{lopez2013evaluating}.
\qald and \kbqa are strictly related, as they both target the retrieval of answers from \ac{KB}s.
However, the current benchmarks and datasets available for evaluating \qald approaches are limited to an \emph{unlinked} and \emph{unstandardized} vision of the structured question answering task.
In this position paper, we point out the lack of existing methods to evaluate \ac{QA} on datasets which exploit external links to the rest of the cloud.
Moreover, we argue that several learning-based \kbqa approaches may be very competitive in \qald challenges, as the current distinctions among their respective benchmarks are only in terms of underlying \ac{KB}s.
Instead, our plea is to let language experts do language and Web semantics experts do semantics.
We propose the creation of new evaluation methods and settings to leverage the advantages of the \ac{SW} to achieve AI-complete \ac{QA} over the Web of Data~\cite{diefenbach2018}.

\section{State of the Art} % line might be commented out if we need space

\subsection{Question Answering over Linked Data}

The most popular datasets for \qald are collected in the homonym \qald benchmarks~\cite{lopez2013evaluating}, which have been released through 9 challenge editions since 2011, the \bioasq challenge for biomedical QA~\cite{tsatsaronis2012bioasq}, and the more recent \lcquad~\cite{trivedi2017lc}. %, a version of Stanford's \textsc{Squad} for structured QA.
18 out of 29 datasets in the aforementioned benchmarks are \emph{open-domain} (i.e., they target a knowledge graph such as DBpedia and Wikidata); 8 \emph{domain-specific} datasets are from the biomedical domain, 3 describe music and 1 describes governmental data.
\ac{QA} systems are supposed to build queries aimed at retrieving information from the RDF dataset itself and, in 3 cases, associated textual resources such as abstracts~\cite{hoffner2017survey}. As \cite{diefenbach2017} reports, the \qald benchmark has been the evaluation subject of diverse systems, some of them based on query templates.
Techniques adopted include graph search, Hidden Markov Models, structured perceptron and (only recently) deep learning, as well as string similarity, language taxonomies, and distributional semantics.

% \todo[inline]{understand where to put these below}
% Linkability.
Even a 5-star LOD dataset may contain vocabulary which cannot be mapped to any other dataset in the cloud because of its uniqueness (e.g., a specific property of a protein might not exist in open-domain \ac{KB}s). 
% Distributed knowledge means different contexts!
In order to achieve QA over the \ac{SW}, we need to perform QA outside of a single KB by exploiting the advantages of the \ac{SW} (e.g., external links, ontology alignments).

\subsection{Question Answering with Knowledge Base}

Within the \ac{SW} community, \kbqa is not as popular as \qald.
Research in Semantic Parsing, defined as \textit{``the task of converting a natural language utterance to a logical form''}, was often excluded from comparisons with \qald approaches for being irrelevant to RDF~\cite{hoffner2017survey}.
% Only recently, it has been rediscovered in some works~\cite{hakimov2015}.
Datasets based on Freebase~\cite{bollacker2008freebase} such as \wqsp~\cite{berant2013semantic}, \simpleq and \complexq~\cite{bordes2015large} are rarely used in the \qald community, despite their full compatibility with RDF standards.
On the other hand, they are widely adopted in the \kbqa community.
Several works in the field of Computational Linguistics target both closed- and open-domain \kbqa. Early on, \cite{yahya2012natural} proposed a supervised method to translate questions into queries, which however required a lot of training data.
Semantic parsing approaches were later introduced to address this problem by learning the queries out of question-answer pairs, as in~\cite{berant2013semantic}.
\emph{Neural Symbolic Machines} were devised for query induction using recurrent neural networks and reinforcement learning~\cite{liang2016}.
Very recently, \cite{abujabal2017automated} proposed an approach to automate the template generation and in~\cite{abujabal2018never}, an architecture to reach never-ending learning from a small set of question-answer pairs.
Most of current \qald approaches still struggle tackling the issues above, in addition to lexical gap and support for complex operators~\cite{hoffner2017survey}.
Given the task similarities, we expect the aforementioned \kbqa approaches to achieve high scores on \qald.

\section{Where is Linked Data?}

We argue that the open-domain \qald benchmarks are to DBpedia and Wikidata what the \kbqa benchmarks are to Freebase.
Judging strictly from a practical viewpoint, the two areas do not differ in anything except in the underlying data.
Hence the question in the title, \textit{``Where is Linked Data in Question Answering over Linked Data''}?

\begin{figure}[h]
    \centering
    \includegraphics[width=0.8\textwidth]{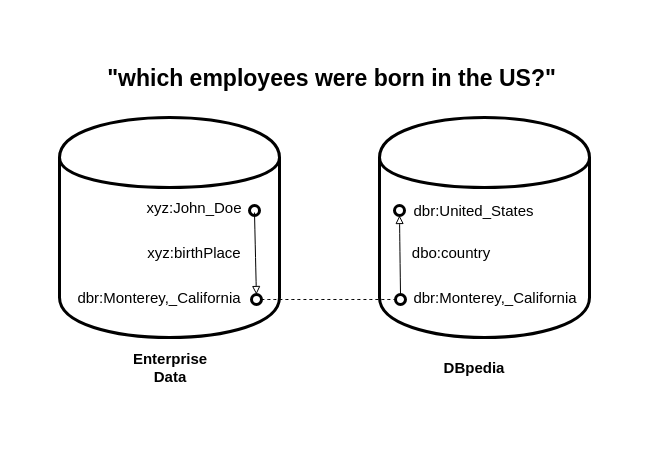}
    \caption{A federated Question Answering example.}
    \label{fig:fqa}
\end{figure}

% We know that the \lod cloud is structurally different from open-domain Knowledge Graphs.
Unfortunately, \qald benchmark datasets are \emph{self-contained}, meaning that the desired information can be either found inside them or not found at all.
Such scenario is different from the \lod cloud, where information about any real-world entity is usually spread across multiple datasets.
As of today, to the best of our knowledge, a method to evaluate QA on datasets which exploits external links to the rest of the cloud does not yet exist.
Let us introduce the example in Figure~\ref{fig:fqa}; say we have a source dataset reporting employee data including the city of their birth.
The question \textit{``which employees were born in the US?''} would need additional knowledge to be answered correctly (i.e., what ``US'' means and how they relate with the cities).
We argue that this kind of QA problems can be addressed only by following the external links from the starting dataset; in this case, an employee can be born in \texttt{dbr:Monterey,\_California} and -- by virtue of the following DBpedia statement -- be added to the result set.
\begin{verbatim}
  dbr:Monterey,_California  dbo:country  dbr:United_States  .
\end{verbatim}
One could argue that there is no conceptual difference in performing QA over two or more interlinked KBs and the same KBs merged into one.
However, in a real-world scenario, Linked Datasets can be extremely different in size, structure, and format, as well as be subject of constant change.
As up-to-dateness does not seem a concern for the \kbqa community -- since Freebase is now a defunct project containing obsolete data -- the \ac{SW} community needs to start considering this problem in its actual environment, i.e. the Web.
% the data and external links in the source dataset are often \emph{unknown} to the user.
% Moreover, the  new links discover more things (thanks, Tim BL!)
% - in the schema, which can be shared and thus enable transfer learning.

Nevertheless, in order to scale to the size of the Web, systems must be prepared.
It is known that scalability is still an issue for the majority of the \qald systems~\cite{diefenbach2017}.
Recent works in \kbqa managed to deal with billions of triples achieving a satisfiable waiting time for the end user~\cite{abujabal2017automated}. 
With this respect, neural approaches are a promising alternative; despite being expensive during training, neural networks usually do not require as many resources for prediction~\cite{sorumarx2017}.

Another point that would differentiate the two research areas is the variety of schemata used in the \lod cloud.
Each dataset has its own vocabulary, which is adapted to the context, where some vocabularies use extremely specific or idiosyncratic terms.
It is therefore not granted that an approach which can perform well on open-domain QA may also generalize well on other datasets.
Term overlap is also frequent, leading to a scenario that is completely different than in \kbqa.

\section{What to do now?}

In this section, we propose two settings of a hypothetical benchmark.
Both settings could be generated semi-automatically without too much human effort, or at least no more than for a canonical \qald dataset.
The availability of the numerous \qald benchmarks released so far reduces the risk factor consistently.
After such new benchmarks are released, we expect the community to react and provide resolutions in the span of one year within the next \qald edition.

\paragraph{Setting 1.}
A domain-specific dataset containing links to (e.g.) DBpedia and Schema.org entities is given in full to the QA systems.
Questions can be answered only if the DBpedia and Schema.org links are dereferenced.
A snapshot of the used \lod cloud subset can be created for reproducibility and distributed through query-ready formats such as KBox~\cite{marx2017kbox} and HDT~\cite{gallego2011hdt}.

\paragraph{Setting 2.}
The second setting we propose exploits the fact that properties utilized in the \lod cloud are defined by widely-adopted \emph{standardized} vocabularies.
The task is to apply transfer learning over two or more datasets using one or more common upper ontologies (e.g., OWL, DCT, SKOS).
Training data are only given on a dataset A, while questions target dataset B, where the different ontologies in A and B are aligned to an upper ontology.

% \todo[inline]{an example, maybe a picture}

\section{Conclusion}

We showed the necessity for the \qald community to tackle the homonym problem not as a single KB, from the perspective of the Web of Data.
After such new benchmarks are released, we expect the community to react and provide resolutions in the span of one year within the next \qald edition.
Our plea is to let language experts do language and Web semantics experts do semantics.
While the former will keep addressing all problems related to human language, new challenges will arise for the \ac{SW}.

\section*{Acknowledgments}

We thank Dennis Diefenbach for his kind suggestions and feedback.

\bibliography{biblio}

\begin{thebibliography}{10}

\bibitem{abujabal2018never}
Abdalghani Abujabal, Rishiraj Saha~Roy, Mohamed Yahya, and Gerhard Weikum.
\newblock Never-ending learning for open-domain question answering over
  knowledge bases.
\newblock In {\em Proc. of the 2018 The Web Conference}, pages 1053--1062,
  2018.

\bibitem{abujabal2017automated}
Abdalghani Abujabal, Mohamed Yahya, Mirek Riedewald, and Gerhard Weikum.
\newblock Automated template generation for question answering over knowledge
  graphs.
\newblock In {\em Proc. of the 26th Int. Conf. on World Wide Web}, pages
  1191--1200, 2017.

\bibitem{berant2013semantic}
Jonathan Berant, Andrew Chou, Roy Frostig, and Percy Liang.
\newblock Semantic parsing on freebase from question-answer pairs.
\newblock In {\em EMNLP}, 2013.

\bibitem{bollacker2008freebase}
Kurt Bollacker, Colin Evans, Praveen Paritosh, Tim Sturge, and Jamie Taylor.
\newblock Freebase: a collaboratively created graph database for structuring
  human knowledge.
\newblock In {\em SIGMOD}, 2008.

\bibitem{bordes2015large}
Antoine Bordes, Nicolas Usunier, Sumit Chopra, and Jason Weston.
\newblock Large-scale simple question answering with memory networks.
\newblock 2015.

\bibitem{diefenbach2018}
Dennis Diefenbach, Andreas Both, Kamal Singh, and Pierre Maret.
\newblock Towards a question answering system over the semantic web.
\newblock {\em arXiv preprint arXiv:1803.00832}, 2018.

\bibitem{diefenbach2017}
Dennis Diefenbach, Vanessa Lopez, Kamal Singh, and Pierre Maret.
\newblock Core techniques of question answering systems over knowledge bases: a
  survey.
\newblock {\em KIS}, 2017.

\bibitem{gallego2011hdt}
Mario~Arias Gallego, Javier~D Fern{\'a}ndez, Miguel~A Mart{\i}nez-Prieto, and
  Claudio Gutierrez.
\newblock Hdt-it: Storing, sharing and visualizing huge rdf datasets.
\newblock In {\em 10th Int. Semantic Web Conference, Bonn, Germany, October},
  pages 23--27, 2011.

\bibitem{hoffner2017survey}
Konrad H{\"o}ffner, Sebastian Walter, Edgard Marx, Ricardo Usbeck, Jens
  Lehmann, and Axel-Cyrille Ngonga~Ngomo.
\newblock Survey on challenges of question answering in the semantic web.
\newblock {\em Semantic Web}, 8(6):895--920, 2017.

\bibitem{liang2016}
Chen Liang, Jonathan Berant, Quoc Le, Kenneth~D Forbus, and Ni~Lao.
\newblock Neural symbolic machines.
\newblock 2016.

\bibitem{lopez2013evaluating}
Vanessa Lopez, Christina Unger, Philipp Cimiano, and Enrico Motta.
\newblock Evaluating question answering over linked data.
\newblock {\em Journal of Web Semantics}, 21:3--13, 2013.

\bibitem{marx2017kbox}
Edgard Marx, Ciro Baron, Tommaso Soru, and S{\"o}ren Auer.
\newblock Kbox — transparently shifting query execution on knowledge graphs
  to the edge.
\newblock In {\em ICSC}, 2017.

\bibitem{mccrae2018lod}
John~P. McCrae, Andrejs Abele, Paul Buitelaar, Richard Cyganiak, Anja Jentzsch,
  and Vladimir Andryushechkin.
\newblock The {L}inked {O}pen {D}ata {C}loud, 2018.

\bibitem{sorumarx2017}
Tommaso Soru, Edgard Marx, Diego Moussallem, Gustavo Publio, Andr\'e
  Valdestilhas, Diego Esteves, and Ciro~Baron Neto.
\newblock {SPARQL} as a foreign language.
\newblock In {\em 13th Int. Conf. on Semantic Systems (SEMANTiCS 2017) -
  Posters and Demos}, 2017.

\bibitem{trivedi2017lc}
Priyansh Trivedi, Gaurav Maheshwari, Mohnish Dubey, and Jens Lehmann.
\newblock Lc-quad: A corpus for complex question answering over knowledge
  graphs.
\newblock In {\em International Semantic Web Conference}, pages 210--218.
  Springer, 2017.

\bibitem{tsatsaronis2012bioasq}
George Tsatsaronis, Michael Schroeder, Georgios Paliouras, Yannis Almirantis,
  et~al.
\newblock Bioasq: A challenge on large-scale biomedical semantic indexing and
  question answering.
\newblock In {\em AAAI fall symposium}, 2012.

\bibitem{yahya2012natural}
Mohamed Yahya, Klaus Berberich, Shady Elbassuoni, Maya Ramanath, Volker Tresp,
  and Gerhard Weikum.
\newblock Natural language questions for the web of data.
\newblock In {\em EMNLP}, pages 379--390. ACL, 2012.

\end{thebibliography}
\bibliographystyle{plain}

\end{document}